\pdfoutput=1

\documentclass[11pt]{article}

\usepackage{emnlp2021}

\usepackage{times}
\usepackage{latexsym}

\usepackage[T1]{fontenc}

\usepackage[utf8]{inputenc}

\usepackage{microtype}
\usepackage{multicol}
\usepackage{multirow}
\usepackage{caption}
\usepackage{graphicx}
\usepackage{amssymb}
\usepackage{amsmath}
\newcommand{\ignore}[1]{}
\usepackage{xspace}
\usepackage{algorithm}
\usepackage{array}
\usepackage{fp}
\usepackage{makecell}
\usepackage{flushend}
\usepackage{cases}
\usepackage{color}
\usepackage{booktabs}
\usepackage{algpseudocode}
\usepackage{listings}
\usepackage{xcolor}
\usepackage{mathrsfs}
\usepackage{subcaption}

\newcommand{\paratitle}[1]{\vspace{1.5ex}\noindent\textbf{#1}}
\newcommand{\ie}{\emph{i.e.,}\xspace}

\newcommand{\eg}{\emph{e.g.,}\xspace}

\title{Virtual Data Augmentation: A Robust and General Framework for Fine-tuning Pre-trained Models}

\author{
\setcounter{footnote}{1}
	Kun Zhou\textsuperscript{\rm{2},\rm{4}}, 
	Wayne Xin Zhao\textsuperscript{\rm{1},\rm{4}}\thanks{$^\dagger$ Corresponding author} ,
	Sirui Wang\textsuperscript{\rm{3}},
	Fuzheng Zhang\textsuperscript{\rm{3}}, \\
	\textbf{Wei Wu}\textsuperscript{\rm{3}} \and
	\textbf{Ji-Rong Wen}\textsuperscript{\rm{1},\rm{2},\rm{4}} \\
	
	\textsuperscript{1}Gaoling School of Artificial Intelligence, Renmin University of China \\
	\textsuperscript{2}School of Information, Renmin University of China. \textsuperscript{3}Meituan Inc., Beijing, China \\
	\textsuperscript{4}Beijing Key Laboratory of Big Data Management and Analysis Methods\\
	\texttt{francis\_kun\_zhou@163.com}, \texttt{batmanfly@gmail.com} \\
	\texttt{\{wangsirui, zhangfuzheng, wuwei30\}@meituan.com},
	\texttt{jrwen@ruc.edu.cn} \\
}

\begin{document}
\maketitle
\begin{abstract}
Recent works have shown that powerful pre-trained language models (PLM) can be fooled by  small perturbations or intentional attacks.
To solve this issue, various data augmentation techniques are proposed to improve the robustness of PLMs.
However, it is still challenging to augment semantically relevant examples with sufficient diversity.
In this work, we present Virtual Data Augmentation (VDA), a general framework for robustly fine-tuning PLMs.
Based on the original token embeddings, we construct a multinomial mixture for augmenting virtual data embeddings, where a masked language model guarantees the semantic relevance and the Gaussian noise provides the augmentation diversity.
Furthermore, a regularized training strategy is proposed to balance  the two aspects.
Extensive experiments on six datasets show that our approach is able to improve the robustness of PLMs and alleviate the performance degradation under adversarial attacks. 
Our codes and data are publicly available at \textcolor{blue}{\url{https://github.com/RUCAIBox/VDA}}.

\ignore{
Although pre-trained language models (PLM) have achieved exciting performance on various NLP tasks, recent works have shown that PLMs can be fooled by adding small perturbations.
To solve this issue, a surge of works adopt data augmentation to improve the robustness.
However, it is challenging to augment diverse and semantic-consistent examples.
In this work, we present Virtual Data Augmentation (VDA), a general framework for robustly fine-tuning PLMs.
We construct a multinomial mixture distribution to aggregate token embeddings for augmenting virtual data embeddings, where a masked language model guarantees the semantic-consistence and a Gaussian noise provides diversity.
Extensive experiments on six datasets show that our method is able to improve the robustness of baseline models and alleviate the performance degradation. 
Our codes and data will be released to support further explorations.}
\end{abstract}

\section{Introduction}
Recently, pre-trained language models (PLMs) such as BERT~\cite{DBLP:conf/naacl/DevlinCLT19} and RoBERTa~\cite{DBLP:journals/corr/abs-1907-11692} have achieved remarkable success in various natural language processing (NLP) tasks~\cite{DBLP:conf/emnlp/RajpurkarZLL16,wang2018glue,zhou2020towards}. 
As a general and effective approach, fine-tuning PLMs on specific datasets has become the mainstream paradigm for developing NLP applications.
Despite the success, researchers have found that PLMs can be easily fooled by adversarial attacks~\cite{DBLP:conf/aaai/JinJZS20,DBLP:conf/emnlp/LiMGXQ20}.
Although encapsulated into a black box, these attack strategies can detect the vulnerabilities of a PLM via intentional queries~\cite{DBLP:journals/corr/abs-2103-10013,DBLP:journals/corr/abs-2009-07502}, and then add small perturbations (\eg synonyms substitution) into the input texts for misleading PLMs to incorrect predictions.

As found in previous works~\cite{DBLP:conf/nips/SchmidtSTTM18,DBLP:conf/icml/YinRB19,DBLP:conf/acl/JiangHCLGZ20}, a possible reason of the vulnerability is that these PLMs do not generalize well on semantic neighborhood around each example in the representation space.
To solve this issue, adversarial data augmentation (ADA) methods~\cite{DBLP:conf/emnlp/JiaL17,DBLP:conf/naacl/WangB18,DBLP:conf/naacl/MichelLNP19} have been proposed by revising original data to augment attack-related data for training.
However, due to the discrete nature of language, it is challenging to generate \emph{semantically relevant}  and \emph{sufficiently diverse}  augmentations.
Although attempts by leveraging expert knowledge~\cite{DBLP:conf/acl/RenDHC19,DBLP:conf/ndss/LiJDLW19} and victim models~\cite{DBLP:conf/aaai/JinJZS20,DBLP:conf/emnlp/LiMGXQ20} have achieved better performance, their generalizability and flexibility is highly limited. 


Recently, virtual adversarial training~\cite{DBLP:conf/iclr/MiyatoDG17,DBLP:conf/iclr/MadryMSTV18} is applied to various NLP models for improving the performance and robustness~\cite{DBLP:conf/iclr/ZhuCGSGL20,DBLP:conf/acl/JiangHCLGZ20}, which usually generates gradient-based perturbation on the embedding space as virtual adversarial samples.
However, it is hard to explicitly constrain the gradient-based perturbation within the same semantic space as the original sample.
In addition, unlike attacks in computer vision~\cite{DBLP:conf/cvpr/ZhengSLG16,DBLP:journals/pami/MiyatoMKI19}, textual adversarial attacks are discrete (\eg word replacement) and are hard to be captured by gradient-based perturbations.

\begin{figure}[t]
\centering
\includegraphics[width=\linewidth]{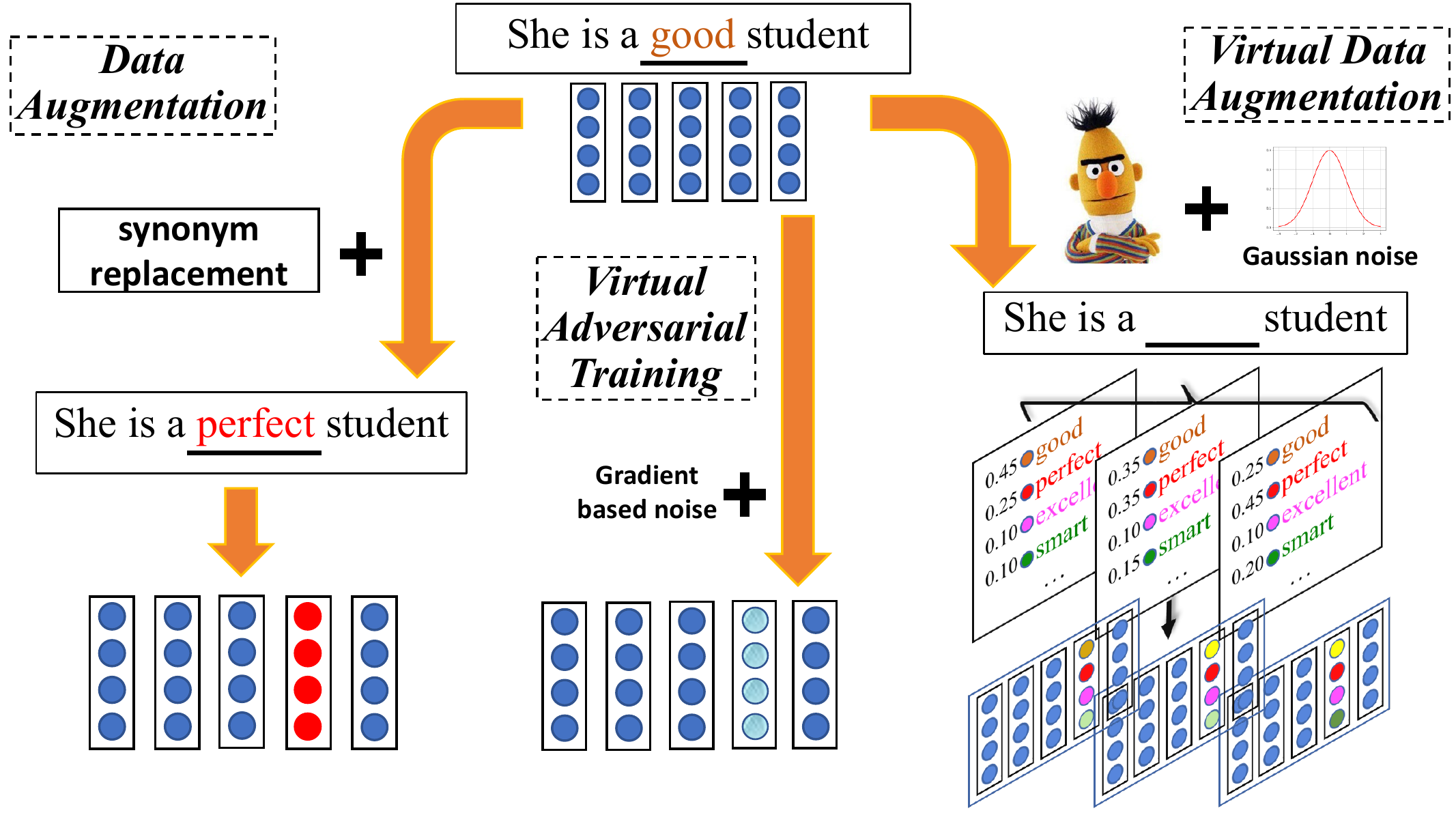}
\caption{A comparison among synonym replacement based data augmentation methods, virtual adversarial training and our VDA framework.}
\label{fig-intro}
\end{figure}

To solve these challenges, we propose Virtual Data Augmentation (VDA), a robust and general framework for fine-tuning pre-trained models.
Our idea is to generate data augmentations at the embedding layer of PLMs. To guarantee semantic relevance, we consider a multinomial mixture of the original token embeddings as the augmented embedding for each position of the input. In the mixture, each token embedding is weighted according to its likelihood estimated by a masked language model conditioned on the input. To provide sufficient diversity, we further incorporate Gaussian noise in the above multinomial mixture, which enhances the randomness of the augmentations. 
As shown in Figure~\ref{fig-intro}, for a target token \emph{``good''}, we first predict the substitution probabilities of candidate tokens via a masked language model, then inject the Gaussian noise to produce multiple multinomial mixtures.
After that, we aggregate the candidate embeddings with the multinomial mixtures to generate new embeddings (virtual data embeddings) to replace the original embedding of \emph{``good''}.

There are two major advantages to our VDA approach. First, with the original token embeddings as the representation basis, the augmented embeddings stay close to the existing embeddings, which avoids the unexpected drift of semantic space. Second, with the injected Gaussian noise, we are able to generate diverse variations for augmentations. In order to enhance the relevance with the given injected Gaussian noise, we further design a regularized training strategy that guides the learning of the augmented virtual data towards the original predictions of PLMs. 
In this way, our approach has considered both semantic relevance and sufficient diversity. 
Besides, since VDA only revises the input embeddings, it is agnostic to downstream tasks, model architectures and learning strategies.

To evaluate the effectiveness of our proposed VDA framework, we construct extensive experiments on six datasets.
Results show that VDA can boost the robustness of all the baseline models without performance degradation.
We also find that our approach can be further improved by combining it with traditional adversarial data augmentation.

Our contributions are summarized as follows:

$\bullet$ We propose a new data augmentation framework for resisting discrete adversarial attacks on PLMs, which is general to improve the robustness of various PLMs on downstream tasks. 

$\bullet$ Our approach utilizes a masked language model with Gaussian noise to augment virtual examples for improving the robustness, and also adopts regularized training to further guarantee the semantic relevance and diversity. 

$\bullet$ Extensive experiments on six datasets have demonstrated that the proposed approach is able to effectively  improve the robustness of PLMs, which can be further improved by combining with existing adversarial data augmentation strategies.

\section{Related Work}
We review the related work in the following three aspects.

\subsection{Adversarial Attack in NLP}
Inspired by the success in compute vision~\cite{DBLP:journals/corr/GoodfellowSS14,DBLP:conf/iclr/KurakinGB17a}, adversarial attack in NLP tasks has become an emerging research topic in recent years~\cite{DBLP:conf/sp/GaoLSQ18,DBLP:journals/jmlr/YangCHWJ20,DBLP:conf/acl/ChenRLL20}.
Early works usually adopt heuristic rules to revise the input text for producing adversarial samples, including character modification~\cite{DBLP:conf/acl/EbrahimiRLD18}, synonyms replacement~\cite{DBLP:conf/emnlp/AlzantotSEHSC18}, word insertion or deletion~\cite{DBLP:conf/acl/ZhangZML19}.
However, with the revolution of large-scale PLMs, these attack strategies can be defended~\cite{DBLP:conf/acl/JonesJRL20,DBLP:journals/corr/abs-2103-11441,zhou2020s3} to some extent.
To attack PLMs, TextFooler~\cite{DBLP:conf/aaai/JinJZS20} designs an attack algorithm to revise the input data and queries the PLM several times to find important words for replacement, which greatly reduces the accuracy of BERT.
Following it, recent works~\cite{DBLP:conf/emnlp/LiMGXQ20,DBLP:journals/corr/abs-2103-10013} continuously improve the quality of the adversarial samples and the attack success ratio.
In our approach, we consider improving the robustness of PLMs against these adversarial attack methods via a new fine-tuning framework VDA.

\subsection{Data Augmentation}
Data augmentation has been extensively studied in NLP tasks for improving the robustness~\cite{DBLP:conf/emnlp/WangY15,DBLP:conf/acl/FadaeeBM17a,DBLP:conf/emnlp/WeiZ19}.
Similar to adversarial attack, early works mostly try heuristic rules to revise the input data for augmentation, such as synonym replacement~\cite{DBLP:conf/naacl/WangB18}, grammar induction~\cite{DBLP:conf/acl/MinMDPL20}, word insert and delete~\cite{DBLP:conf/emnlp/WeiZ19}.
With the development of text generation techniques, back translation~\cite{DBLP:conf/nips/XieDHL020,DBLP:conf/acl/SinghGR18} and variant auto-encoder~\cite{DBLP:conf/emnlp/WangPPCWL20,DBLP:conf/aaai/LiQTC0Y19} are used to augment new data.
Besides, a surge of works~\cite{DBLP:conf/coling/HouLCL18,DBLP:conf/emnlp/LiLHZZ19,zhou2019unsupervised} focus on augmentation for specific tasks with special rules or models.
Although they perform well, these methods have lost the generality.
In this paper, we propose a new data augmentation framework VDA that utilizes a masked language model with Gaussian noise to augment virtual examples for improving the robustness. our VDA is agnostic to downstream tasks, model architectures and learning strategies.

\subsection{Virtual Adversarial Training}
To improve the robustness of neural networks against adversarial examples, virtual adversarial training (VAT)~\cite{miyato2015distributional,DBLP:conf/iclr/KurakinGB17a,DBLP:conf/nips/QinMGKDFDSK19} has been widely used in compute vision. It formulates a class of adversarial training algorithms into solving a minimax problem, which can be achieved reliably through multiple projected gradient ascent steps~\cite{DBLP:conf/nips/QinMGKDFDSK19}.
Recently, VAT has shown its effectiveness in NLP tasks, where the gradient-based noise is able to improve the performance and smoothness of the pre-trained models~\cite{DBLP:conf/iclr/ZhuCGSGL20,DBLP:conf/acl/JiangHCLGZ20}.
However, due to the discrete nature of language, it has been shown that VAT methods are not very effective in defending against adversarial attacks~\cite{DBLP:journals/corr/abs-2012-15699,DBLP:conf/aaai2021/tavat}.
\section{Preliminary}
This work seeks to improve the fine-tuning performance of pre-trained language models~(PLM), in that the fine-tuned model will become more robust to data permutations or attacks. Specially, we take the text classification task as an example task to illustrate our approach, where a set of $n$ labeled texts $\{ \langle x_i, y_i \rangle \}$ are available. Each labeled text consists of a text $x_i$ and a label $y_i$ from the label set $\mathcal{Y}$.  We refer to the adversarial example generated from a text $x_i$ as adversarial text, denoted by  $\hat{x}_i$. The purpose of adversarial examples is to enhance the model robustness in resisting intentional data perturbations or attacks. 

Let $f$ denote a PLM parameterized by $\theta$.
Following~\cite{DBLP:conf/emnlp/JiaL17,DBLP:conf/naacl/MichelLNP19}, we incorporate adversarial examples to improve the fine-tuning of PLMs.
To conduct the adversarial learning, we formulate the learning objective as follows:
\begin{equation}\small
\arg\min_{\theta} \sum_{i=1}^n L_{\text{c}}(f(x_i),y_i) + \lambda \sum_{j=1}^{m} L_{\text{reg}}(f(x_j),f(\hat{x}_j)),
\label{eq-origin}
\end{equation}
where $m$ is the number of adversarial texts that we use, $\lambda$ is a trade-off parameter, $L_{\text{c}}$ and $L_{\text{reg}}$ try to minimize the classification loss and reduce the prediction difference between original and adversarial texts, respectively.  

For the PLM $f$, we assume that it is already pre-trained on general-purpose large-scale text data, we would like to fine-tune its parameter $\theta$ based on some downstream task. The PLMs are usually developed based on multi-layered Transformer architecture such as BERT~\cite{DBLP:conf/naacl/DevlinCLT19} and RoBERTa~\cite{DBLP:journals/corr/abs-1907-11692}, where a sequence of tokens will be encoded into a sequence of contextual representations. Here, we take the representation of the first token (\ie \texttt{[CLS]}) as the input of the classifier, and optimize the classification performance with the cross-entropy loss.

\section{Our Approach}
In this section, we describe our proposed framework \emph{Virtual Data Augmentation (VDA)} for robustly fine-tuning PLMs. 
Our framework consists of two important ingredients, namely embedding augmentation and regularized training.

\begin{figure}[t]
\centering
\includegraphics[width=\linewidth]{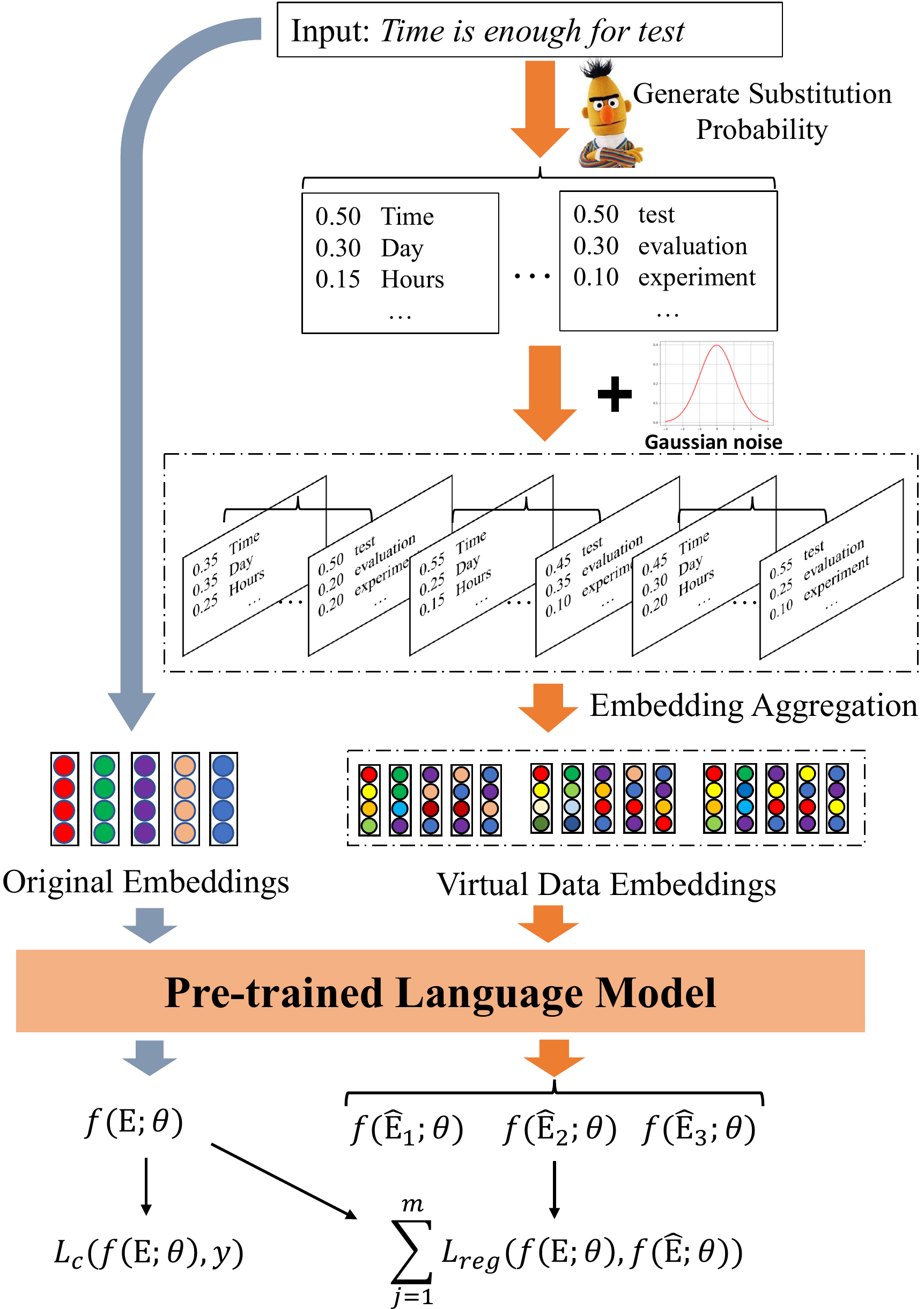}
\caption{Illustration of our framework VDA. We show the case that we generate three virtual examples for the input sentence.}
\label{intro}
\end{figure}

\subsection{Embedding Augmentation}

To improve the model robustness, a good adversarial example should adhere to the original semantic space, as well as incorporate sufficient variations in meanings. However, existing studies cannot make a good trade-off between the two aspects. 

Considering this difficulty, we generate adversarial texts at the embedding layer of PLMs. For adversarial training, continuous embeddings are easier to optimize and can encode more semantic variations than discrete tokens.  
The key idea of embedding augmentation is inspired by the word replacement strategy in previous data augmentation methods~\cite{DBLP:conf/naacl/Kobayashi18,DBLP:conf/emnlp/WeiZ19}. Instead of selecting some tokens for replacement, we use an augmented embedding to replace the original contextual embedding of a specific  token in the input sentence. 
To adhere to the original semantic space, the augmented embedding is derived by a probabilistic mixture of the embeddings of the vocabulary terms, where each term is weighted according to its \emph{substitution probability} (\ie replacing the original token with the candidate term)  calculated by a masked language model (MLM). 



To simplify our presentation, we only discuss the augmentation for a specific token $\tilde{w}$ from an input sentence $S$. The same procedure will be applied to each position of the original sentence $S$. 
Specially, we utilize the MLM to evaluate the substitution probabilities of all the terms in the vocabulary. For each chosen token, we predict its probability to be replaced by other words in the whole vocabulary via MLM, denoted as $p(\hat{w}_{i}|S)$.
Finally, we obtain the substitution probabilities of all the terms as 
\begin{equation}
\{p(\hat{w}_{1}|S), \dots, p(\hat{w}_{V}|S)\},
\end{equation}
where $V$ is the vocabulary size. 
Different from previous masked prediction~\cite{DBLP:conf/naacl/DevlinCLT19}, we do not mask the chosen token but also keep it as the input to compute the substitution probabilities. 
In this way, we aim to generate very relevant embeddings for augmentation. 
Such a strategy is also very efficient in practice, since it no longer performs the costly mask-and-completion operations for each token.

To augment diverse virtual data embeddings, we further draw a random noise from the Gaussian distribution as
\begin{equation}
    \epsilon \sim \mathcal{N}(0, \sigma^{2}),
\end{equation}
where the randomness can be controlled by the standard variance $\sigma$. 
By mixing the random noise with the substitution probabilities, we can produce multiple different probability distributions for each instance as
\begin{equation}
    p^{'}(\hat{w}_{i}|S)=\texttt{softmax}(p(\hat{w}_{i}|S)+\epsilon).
    \label{eq-noise}
\end{equation}
Then, for each target token $\tilde{w}$, we obtain its corresponding substituted embedding by aggregating the token embedding matrix according to the noised substitution probability as
\begin{equation}
    \hat{\mathbf{e}}_{\tilde{w}}=\mathbf{p}_{\tilde{w}} \cdot \mathbf{M}_{E},
    \label{eq-emb}
\end{equation}
where $\mathbf{p}_{\tilde{w}}=\{ p^{'}(\hat{w}_{i}|S) \}_{i=1}^{V}$, and $\mathbf{M}_{E}\in \mathbb{R}^{V \times d}$ is the token embedding matrix from the MLM.
Note that by using the output of MLM, our approach can augment more ``real'' embeddings from the  semantic space spanned by original token embeddings.
Besides, mixing Gaussian noise brings additional semantic diversity for augmentation. 

\subsection{Regularized Training}
The above augmentation strategy is able to enhance the semantic variations by continuous embeddings. However, augmented data is likely to incorporate  unexpected semantic drift in representations. 
To further improve the model robustness, instead of directly using the augmented embeddings as positive examples, we propose a regularized training strategy to prevent large changes between the predictions given real and augmented embeddings. Formally, given the original data point $(\mathbf{E}_i, y_i)$ and the augmented virtual data $\hat{\mathbf{E}}_i$, where $\mathbf{E}_i$ and $\hat{\mathbf{E}}_i$ denote the original embeddings and augmented embeddings of the instance respectively, we set the regularization loss in Equation~\ref{eq-origin}:
\begin{equation}
    L_{\text{reg}}(\theta)=\frac{1}{k}\sum_{i=1}^{k}D_{sKL}\bigg(f(\mathbf{E}_i; \theta), f(\hat{\mathbf{E}}_i; \theta)\bigg),
\end{equation}
where $D_{sKL}$ is the symmetric KL-divergence, $k$ denotes the number of augmented examples.
The regularizer enforces the model $f$ to produce similar scores for the original data and augmented data, which lies in the semantic neighborhood of original embeddings.
Furthermore, we instantiate the classification loss in Equation~\ref{eq-origin} as follows: 
\begin{equation}
    L_c(\theta)=\frac{1}{n}\sum_{i=1}^{n}\text{CE}\bigg(f(\mathbf{E}_i; \theta), y_i\bigg),
\end{equation}
where $CE(\cdot, \cdot)$ is the cross-entropy loss function, which can be changed according to specific tasks. 

\subsection{Overview and Discussion}
\begin{algorithm}[tbp]
\small
	\begin{algorithmic}[1]
		\Require Pre-trained model $f(.)$, a pre-trained masked language model $f_{MLM}(.)$, Gaussian distribution $\mathcal{N}(0, \sigma^{2})$, virtual data sampling number $k$, training epoch $m$.
		\State \textbf{Input:} The training data $\{ \langle x_i, y_i \rangle \}$.
		\State \textbf{Output:} The fine-tuned parameters $\theta$.
		\For {$i=1$ \dots $m$}
		    \For {minibatch $B\in \{ \langle x_i, y_i \rangle \}$}
		        \State Tokenize input sentences in $B$ into $\{w_{1}, ..., w_{m}\}$.
		        \State Generate the substitution probability of all tokens.
		        \For {$j=1$ \dots $k$}
		            \State Sample $\epsilon$ from $N(0, \sigma^{2})$.
		            \State Produce $p^{'}(\hat{w}_{i}|S)$ using Eq.~\ref{eq-noise}.
		            \State Augment virtual data using Eq.~\ref{eq-emb}.
		            \State Optimize $\theta$ using Eq.~\ref{eq-origin}.
		        \EndFor
		    \EndFor
		\EndFor
		\State \Return $\theta$
	\end{algorithmic}
	\caption{The framework of VDA.}
	\label{alg-vda}
\end{algorithm}
In this part, we present the overview and discussions of our VDA approach.

\paratitle{Overview}
The overall framework of VDA consists of two important parts, namely embedding augmentation and regularized training. 
We present the overall training algorithm in Algorithm~\ref{alg-vda}.
For embedding augmentation, we utilize the output of a MLM as the multinomial mixtures to augment new embeddings for each token in the input sentence.
It is called \emph{virtual data augmentation}, since the augmented embeddings do not correspond to actual text or tokens, but a probabilistic mixture of all the token embeddings.
Then, for regularized training, we leverage the original predictions to guide the learning of the augmented embeddings,  which reduces the influence from noisy or incorrect perturbations in the augmentations.

\paratitle{Discussion}
In the background of machine learning~\cite{DBLP:conf/nips/SchmidtSTTM18,DBLP:conf/icml/YinRB19}, 
robustness corresponds to the ability to resist data drift, perturbation and attack.
To improve the robustness, a key point is that the model is able to generalize to the semantic neighborhood of training data instances~\cite{DBLP:conf/nips/SchmidtSTTM18}. 
However, discrete augmentation methods~\cite{DBLP:conf/emnlp/WeiZ19,DBLP:conf/naacl/WangB18} (\eg insert, delete or replace tokens) do not have good generalization ability for model optimization. While, virtual adversarial training methods~\cite{DBLP:conf/iclr/ZhuCGSGL20,DBLP:conf/acl/JiangHCLGZ20} cannot well constrain the augmentations in the original semantic space. As a comparison, our approach utilizes original token embeddings to augment new embeddings, so that the augmentations will stay close to the existing  embeddings in the same semantic space. For relevance, we adopt a MLM to generate  the multinomial mixture according to the likelihood of each candidate given the input. For diversity, we inject Gaussian noise to enhance the randomness. To further balance the two aspects, we design a regularized strategy to guide the augmentation learning towards the original predictions. By only revising the embeddings, our approach is model-agnostic and domain-agnostic, which is general to apply to various PLMs on different downstream tasks.   

\section{Experiment - Main Results}
We demonstrate the effectiveness of VDA for fine-tuning PLMs in the text classification task.
\begin{table}[t]
\centering
\small
\begin{tabular}{|p{21mm}<{\centering}|l|ccc|}
\hline
\textbf{Task} & \textbf{Data} & \textbf{Train} & \textbf{Dev} & \textbf{Test} \\
\hline
\multirow{4}*{\shortstack{\textbf{Sentence} \\\textbf{Classification}}} & Yelp & 25000 & 5000 & 5000 \\
& IMDB & 25000 & 5000 & 5000 \\
& AG & 120000 & 3000 & 3000 \\
& MR & 8595 & 1000 & 1000 \\
\hline
\multirow{2}*{\shortstack{\textbf{Sentence Pair} \\\textbf{Classification}}} & QNLI & 100000 & 4743 & 5463 \\
& MRPC & 3668 & 408 & 1725 \\ 
\hline
\end{tabular}
\caption{Statistics of the datasets.}
\label{tab-static}
\end{table}

\subsection{Experimental Setup}
\subsubsection{Dataset}

\begin{table*}[t]
\centering
\begin{tabular}{|l|cccc|cccc|}
\hline
\textbf{Datasets} & \multicolumn{4}{|c|}{\textbf{Yelp}} & \multicolumn{4}{|c|}{\textbf{IMDB}} \\
\hline
Metrics & Ori Acc & Att Acc & Q Num & Perturb & Ori Acc & Att Acc & Q Num & Perturb \\
\hline
BERT & 0.957 & 0.487 & 645.878 & 12.291 & \textbf{0.938} & \textbf{0.310} & 954.852 & 9.745 \\
BERT$_{VDA}$ & \textbf{0.959} & \textbf{0.533} & \textbf{687.681} & \textbf{17.720} & \textbf{0.938} & 0.307 & \textbf{1061.656} & \textbf{12.483} \\
\hline
FreeLB & 0.960 & 0.400 & 604.952 & 17.559 & 0.936 & 0.137 & 784.833 & 10.727 \\
FreeLB$_{VDA}$ & \textbf{0.962} & \textbf{0.507} & \textbf{694.910} & \textbf{18.869} & \textbf{0.939} & \textbf{0.250} & \textbf{1015.518} &  \textbf{14.400} \\
\hline
SMART & 0.960 & 0.437 & 636.523 & 20.356 & \textbf{0.940} & 0.110 & \textbf{611.279} & 9.232 \\
SMART$_{VDA}$ & \textbf{0.962} & \textbf{0.527} & \textbf{691.941} & \textbf{20.372} & 0.938 & \textbf{0.143} & \textbf{824.192} & \textbf{12.785} \\
\hline
SMix & 0.957 & 0.557 & 705.042 & 15.305 & \textbf{0.935} & 0.210 & 833.642 & 9.816 \\
SMix$_{VDA}$ & \textbf{0.960} & \textbf{0.600} & \textbf{765.517} & \textbf{17.491} & 0.934 & \textbf{0.383} & \textbf{1105.819} & \textbf{12.733} \\
\hline
RoBERTa & \textbf{0.977} & 0.500 & 730.741 & \textbf{11.991} & 0.957 & \textbf{0.233} & 911.250 & 9.588 \\
RoBERTa$_{VDA}$ & 0.972 & \textbf{0.643} & \textbf{780.300} & 9.323 & \textbf{0.959} & 0.210 & \textbf{960.996} & \textbf{11.469} \\
\hline
\hline
\textbf{Datasets} & \multicolumn{4}{|c|}{\textbf{AG}} & \multicolumn{4}{|c|}{\textbf{MR}} \\
\hline
Metrics & Ori Acc & Att Acc & Q Num & Perturb & Ori Acc & Att Acc & Q Num & Perturb \\
\hline
BERT & 0.944 & 0.327 & 223.271 & \textbf{22.309} & 0.868 & 0.210 & 58.217 & 20.291 \\
BERT$_{VDA}$ & \textbf{0.946} & \textbf{0.450} & \textbf{268.981} & 21.757 & \textbf{0.878} & \textbf{0.339} & \textbf{70.519} & \textbf{22.775} \\
\hline
FreeLB & \textbf{0.945} & 0.301 & 215.178 & 21.175 & 0.879 & 0.240 & 60.498 & 20.266 \\
FreeLB$_{VDA}$ & \textbf{0.945} & \textbf{0.473} & \textbf{271.282} & \textbf{22.634} & \textbf{0.883} & \textbf{0.302} & \textbf{69.109} & \textbf{21.557} \\
\hline
SMART & 0.944 & 0.403 & 251.643 & 21.953 & 0.880 & 0.226 & 56.067 & 20.514 \\
SMART$_{VDA}$ & \textbf{0.945} & \textbf{0.484} & \textbf{273.724} & \textbf{23.572} & \textbf{0.885} & \textbf{0.298} & \textbf{63.965} &  \textbf{23.386} \\
\hline
SMix & 0.944 & 0.425 & 269.616 & 21.981 & 0.880 & 0.251 & 61.289 & \textbf{21.842} \\
SMix$_{VDA}$ & \textbf{0.947} & \textbf{0.513} & \textbf{278.833} & \textbf{24.422} & \textbf{0.883} & \textbf{0.319} & \textbf{68.893} & 20.874 \\
\hline
RoBERTa & 0.951 & 0.464 & 301.749 & 19.716 & 0.919 & 0.344 & 81.432 & \textbf{27.128} \\
RoBERTa$_{VDA}$ & \textbf{0.952} & \textbf{0.497} & \textbf{303.964} & \textbf{23.127} & \textbf{0.925} & \textbf{0.439} & \textbf{89.427} & 24.310 \\
\hline
\end{tabular}
\caption{Main results on the sentence classification task. Ori Acc, Att acc, Q Num and Perturb denote the original accuracy, attack accuracy, query number and perturbed percentage per sample. ``$_{VDA}$'' denotes that the model is trained with our proposed VDA framework. The best results in each group are highlighted in bold.}
\label{tab-main-single}
\end{table*}

\begin{table*}
\centering
\begin{tabular}{|l|cccc|cccc|}
\hline
\textbf{Datasets} & \multicolumn{4}{|c|}{\textbf{MRPC}} & \multicolumn{4}{|c|}{\textbf{QNLI}} \\
\hline
\hline
Metrics & Ori Acc & Att Acc & Q Num & Perturb & Ori Acc & Att Acc & Q Num & Perturb \\
\hline
BERT & 0.826 & 0.163 & 77.276 & 9.601 & 0.909 & 0.342 & 93.515 & 13.451 \\
BERT$_{VDA}$ & \textbf{0.831} & \textbf{0.206} & \textbf{90.686} & \textbf{10.617} & \textbf{0.913} & \textbf{0.410} & \textbf{112.088} & \textbf{14.816} \\
\hline
FreeLB & 0.827 & 0.154 & 82.372 & 10.193 & 0.910 & 0.363 & 98.703 & 14.283 \\
FreeLB$_{VDA}$ & \textbf{0.838} & \textbf{0.205} & \textbf{87.379} & \textbf{10.566} & \textbf{0.915} & \textbf{0.428} & \textbf{111.812} & \textbf{15.953} \\
\hline
SMART & 0.831 & 0.139 & 81.114 & 10.270 & 0.909 & 0.309 & 91.435 & \textbf{14.654} \\
SMART$_{VDA}$ & \textbf{0.832} & \textbf{0.179} & \textbf{85.628} & \textbf{11.546} & \textbf{0.911} & \textbf{0.388} & \textbf{105.918} & 14.611 \\
\hline
SMix & 0.824 & 0.249 & \textbf{116.660} & 11.224 & 0.886 & 0.171 & 71.849 & 10.171 \\
SMix$_{VDA}$ & \textbf{0.833} & \textbf{0.258} & 97.380 & \textbf{11.448} & \textbf{0.915} & \textbf{0.389} & \textbf{109.562} & \textbf{14.537} \\
\hline
RoBERTa & 0.850 & 0.179 & 78.905 & 9.357 & 0.934 & 0.408 & 101.977 & \textbf{12.112} \\
RoBERTa$_{VDA}$ & \textbf{0.859} & \textbf{0.255} & \textbf{97.485} & \textbf{11.623} & \textbf{0.941} & \textbf{0.411} & \textbf{103.405} & 11.901 \\
\hline
\end{tabular}
\caption{Main results on the sentence-pair classification task. ``$_{VDA}$'' denotes that the model is trained with our proposed VDA framework. The best results in each group are highlighted in bold.}
\label{tab-pair}
\end{table*}

We conduct experiments on the sentence classification task and the sentence-pair classification task. The dataset statistics are summarized in Table~\ref{tab-static}.

\paratitle{Sentence Classification} We use four sentence classification datasets for evaluation.

$\bullet$ \textbf{Yelp}~\cite{DBLP:conf/nips/ZhangZL15}~\footnote{https://www.yelp.com/}: a binary sentiment classification dataset based on restaurant reviews.

$\bullet$ \textbf{IMDB}~\footnote{https://datasets.imdbws.com/}: a binary document-level sentiment classification dataset on movie reviews.

$\bullet$ \textbf{AG's News}~\cite{DBLP:conf/nips/ZhangZL15}: a news-type classification dataset, containing 4 types of news: World, Sports, Business, and Science.

$\bullet$ \textbf{MR}~\cite{DBLP:conf/acl/PangL05}: a binary sentiment classification dataset based on movie reviews.

\paratitle{Sentence-Pair Classification} We also use two sentence-pair classification datasets for evaluation. 

$\bullet$ \textbf{QNLI}~\cite{DBLP:journals/corr/abs-1809-02922}: a question-answering dataset consisting of question-paragraph pairs. The task is to determine whether the context sentence contains the answer to the question. 

$\bullet$ \textbf{MRPC}~\cite{DBLP:conf/acl-iwp/DolanB05}: a corpus of sentence pairs with human annotations about the semantic equivalence.

\subsubsection{Baselines}
To evaluate the generalization of our framework, we implement VDA on the following models.

$\bullet$ \textbf{BERT-Base}~\cite{DBLP:conf/naacl/DevlinCLT19} is the 12-layer BERT model with 768 hidden units and 12 heads, totally 110M parameters. 

$\bullet$ \textbf{FreeLB}~\cite{DBLP:conf/iclr/ZhuCGSGL20} is an adversarial training approach for fine-tuning PLMs,  which adds gradient-based perturbations to token embeddings. We implement it on BERT-Base.

$\bullet$ \textbf{SMART}~\cite{DBLP:conf/acl/JiangHCLGZ20} is a robust and efficient computation framework for fine-tuning PLMs. Limited by the GPU resource, we can only implement the smooth-inducing adversarial regularization on BERT-Base but remove the Bregman Proximal Point Optimization.

$\bullet$ \textbf{SMix}~\cite{DBLP:journals/corr/abs-2012-15699} uses mixup on [CLS] tokens of the PLM to cover larger attack space. We implement it on BERT-Base. For a fair comparison, we remove the adversarial data augmentation strategy here, and leave it on Section~\ref{sec-ada}.

$\bullet$ \textbf{RoBERTa-Large}~\cite{DBLP:journals/corr/abs-1907-11692} is a robustly optimized BERT model with more training data and time. It owns 24 layers, 1024 hidden units and 16 heads, totally 355M parameters. 

\subsubsection{Evaluation Metrics}
We set up various metrics for measuring accuracy and robustness.
Original accuracy, is the accuracy of models on the original test set. While attack accuracy is the counter-part of after attack accuracy, which is the core metric measuring the robustness.
Larger attack accuracy reflects better robustness.
In this paper, we adopt BERT-Attack~\cite{DBLP:conf/emnlp/LiMGXQ20} as the attack method, since it can generate fluent and semantically preserved samples. 
For AG, MR, QNLI and MRPC datasets, we follow previous works~\cite{DBLP:conf/aaai/JinJZS20,DBLP:conf/emnlp/LiMGXQ20} to randomly sample 1000 instances for robustness evaluation. 
For Yelp and IMDB, we randomly sample 300 instances since the long sentences in the two datasets are more time-consuming.
Note that for sentence-pair classification datasets (\ie QNLI and MRPC), we attack the second sentence in evaluation.
Besides, we also apply the query number and perturbed percentage per sample for evaluation.
Under the black-box setting, queries of the target model are the only way of attack methods to access information. The larger query number indicates that the vulnerability of the target model is harder to be detected, which reflects better robustness.
The perturbed percentage is the ratio of perturbed words number to the text length, a larger percentage also reveals more difficulty to successfully attack the model.

\subsubsection{Implementation Details}
We implement all baseline models based on HuggingFace Transformers~\footnote{https://huggingface.co/transformers/}, and their hyper-parameters are set following the suggestions from the original papers.
For our proposed VDA, we reuse the same hyper-parameter setting as the original baseline model.
All models are trained on a GeForce RTX 3090.

For hyper-parameters in VDA, the sampling number $m$ is set as 1, the learning rate is 1e$^{-5}$. We use 5\% steps to warm up PLMs during training.
The variance of Gaussian noise is mostly set as 1e$^{-2}$ and tuned in $\{1e^{-3}, 4e^{-3}, 1e^{-2}, 4e^{-2}\}$, the weight $\lambda$ is mostly set as 1.0 and tuned in $\{0.04, 0.1, 0.4, 1.0, 4.0\}$.

\subsection{Main Results}

Table~\ref{tab-main-single} reports the evaluation results of our proposed VDA framework and the baseline models on sentence classification datasets. And the results on sentence-pair classification datasets are shown in Table~\ref{tab-pair}. 
Based on these results, we can find:

First, FreeLB and SMART mostly outperform BERT-base model on the original accuracy metric, but perform not well on robustness-related metrics, especially on Yelp and IMDB datasets.
These methods adopt gradient-based perturbations and smoothness-inducing regularization, respectively, which are able to improve the classification accuracy but may be not effective in defending against adversarial attacks.
A potential reason may be that textual adversarial attacks are discrete, which can not be captured by virtual adversarial training.

Second, SMix improves the robustness of BERT-base in all datasets, but performs not well in original accuracy.
It mixes hidden representations of the BERT-base model, which increases the coverage of the attack space for PLMs but may augment noised examples into training data.
Besides, RoBERTa-large outperforms all other baselines in performance and robustness metrics. The reason is that RoBERTa-large is pre-trained on more training data with more training time, which can directly improve the generalization and robustness to adversarial attack samples.

Finally, we compare our proposed framework with these baseline models.
After being combined with VDA, it is clear to see a significant improvement in robustness metrics on most of datasets. 
Our VDA utilizes a masked language model to generate substitution probabilities, and then add a Gaussian noise.
In this way, we can augment diverse and semantic-consistent examples, which are able to improve the robustness of PLMs.
Furthermore, we can also see that the most of baseline models combined with VDA achieve a marginal improvement in original accuracy.
It indicates that our approach can better balance the performance and robustness of PLMs.
Among them, we can see that our VDA can bring more improvement in MRPC and QNLI.
The reason may be that the two tasks are more difficult and require more data for training.
The virtual augmented data via our approach is semantic-consistent and diverse, hence it can be more helpful for these tasks.

\section{Experiment - Analysis and Extension}
In this section, we continue to study and analyze the effectiveness of our proposed VDA.

\subsection{Ablation and Variation Study}
\begin{table}
\small
\begin{tabular}{|l|cccc|}
\hline
& \multicolumn{4}{|c|}{\textbf{AG}} \\
\hline
Method & Ori Acc & Att Acc & Q Num & Perturb \\
\hline
BERT & 0.944 & 0.360 & 241.758 & 22.416  \\
+${VDA}$ & \textbf{0.949} & 0.468 & 284.946 & 22.470 \\
\hline
+${VDA - \epsilon}$ & 0.945 & 0.445 & 280.672 & 21.642 \\
+${CEVDA}$ & 0.945 & 0.451 & 275.034 & 20.621 \\
+${Argmax}$ & 0.943 & \textbf{0.478} & \textbf{298.186} & 19.515\\
+${Sample}$ & 0.946 & 0.459 & 274.553 & \textbf{22.479}\\
\hline
\hline
& \multicolumn{4}{|c|}{\textbf{QNLI}} \\
\hline
Method & Ori Acc & Att Acc & Q Num & Perturb \\
\hline
BERT & 0.834 & 0.163 & 77.276 & 9.601  \\
+${VDA}$ & 0.838 & \textbf{0.206} & \textbf{90.686} & 10.617 \\
\hline
+${VDA - \epsilon}$ & 0.833 & 0.184 & 86.914 & \textbf{11.075} \\
+${CEVDA}$ & 0.832 & 0.173 & 80.389 & 10.492 \\
+${Argmax}$ & 0.834 & 0.160 & 71.074 & 9.224 \\
+${Sample}$ & \textbf{0.840} & 0.149 & 74.197 & 9.263 \\
\hline
\end{tabular}
\caption{Ablation and variation study of our approach on the developed set of AG and QNLI datasets. BERT indicates the BERT-base model.}
\label{table-ab}
\end{table}

We devise four variations for exploring the effectiveness of key components in our proposed VDA.
BERT$+VDA-\epsilon$ is the variation by removing the Gaussian noise $\epsilon$ in Eq.~\ref{eq-noise}. BERT$+CEVDA$ replaces the symmetric KL-divergence by cross-entropy loss.
BERT+$Argmax$ and BERT+$Sample$ adopt argmax and sample operators to select the substituted token according to the substitution probability, respectively.
We conduct the experiments on AG and QNLI datasets.

As shown in Table~\ref{table-ab}, most of the variations perform better than BERT in robustness metrics, since they all augment virtual data for improving the robustness.
Among them, BERT+$VDA$ outperforms most of the variations in both accuracy and robustness metrics.
It indicates that the Gaussian noise, symmetric KL-divergence loss and weighted aggregated embeddings are all useful to improve the robustness and stabilize the accuracy.
However, we can see BERT+$Argmax$ and BERT+$Sample$ achieve better results than BERT+$VDA$ in part of metrics, but cause a dramatic drop in other metrics.
It indicates that the two variations can not balance the trade-off between accuracy and robustness well. 

\subsection{Virtual Data Augmentation with Adversarial Data Augmentation}
\label{sec-ada}
\begin{table}
\small
\begin{tabular}{|l|cccc|}
\hline
& \multicolumn{4}{|c|}{\textbf{MR}} \\
\hline
Metrics & Ori Acc & Att Acc & Q Num & Perturb \\
\hline
BERT-base & 0.866 & 0.215 & 61.886 & 21.243  \\
+${VDA}$ & \textbf{0.874} & 0.326 & 70.681 & 21.577 \\
+${ADA}$ & 0.862 & 0.287 & 72.635 & \textbf{22.627} \\
+$VDA$+$ADA$ & 0.869 & \textbf{0.386} & \textbf{77.586} & 20.781 \\
\hline
\hline
& \multicolumn{4}{|c|}{\textbf{MRPC}} \\
\hline
Metrics & Ori Acc & Att Acc & Q Num & Perturb \\
\hline
BERT-base & 0.834 & 0.163 & 77.276 & 9.601  \\
+${VDA}$ & \textbf{0.838} & 0.206 & 90.686 & 10.617 \\
+${ADA}$ & 0.828 & 0.214 & 89.721 & 10.317 \\
+$VDA$+$ADA$ & 0.837 & \textbf{0.215} & \textbf{95.786} & \textbf{10.623} \\
\hline
\end{tabular}
\caption{The study of combining VDA and ADA on the developed set of MR and MRPC datasets, BERT indicates the BERT-base model.}
\label{table-ada}
\end{table}

Our proposed VDA is general to various methods, including conventional adversarial data augmentation (ADA).
In this part, we collect the adversarial examples curated from the MR and MRPC training sets, and add them to the original training set, respectively.
Then we test the accuracy and the robustness of BERT-base model and our VDA after training with the adversarial data.
As seen in Table~\ref{table-ada}, although augmented adversarial data improves the robustness of BERT, the performance on original accuracy also drops.
The reason may be that there are noised instances in the adversarial data.
As a comparison, our proposed VDA can augment diverse and semantic-consistent virtual data, which better balances accuracy and robustness.
Besides, after combining with ADA, our VDA can be further improved on accuracy and robustness metrics.
It indicates that our approach is also general to ADA methods.

\subsection{Hyper-parameter Analysis}
\ignore{
\begin{figure}[t]
\centering
\includegraphics[width=\linewidth]{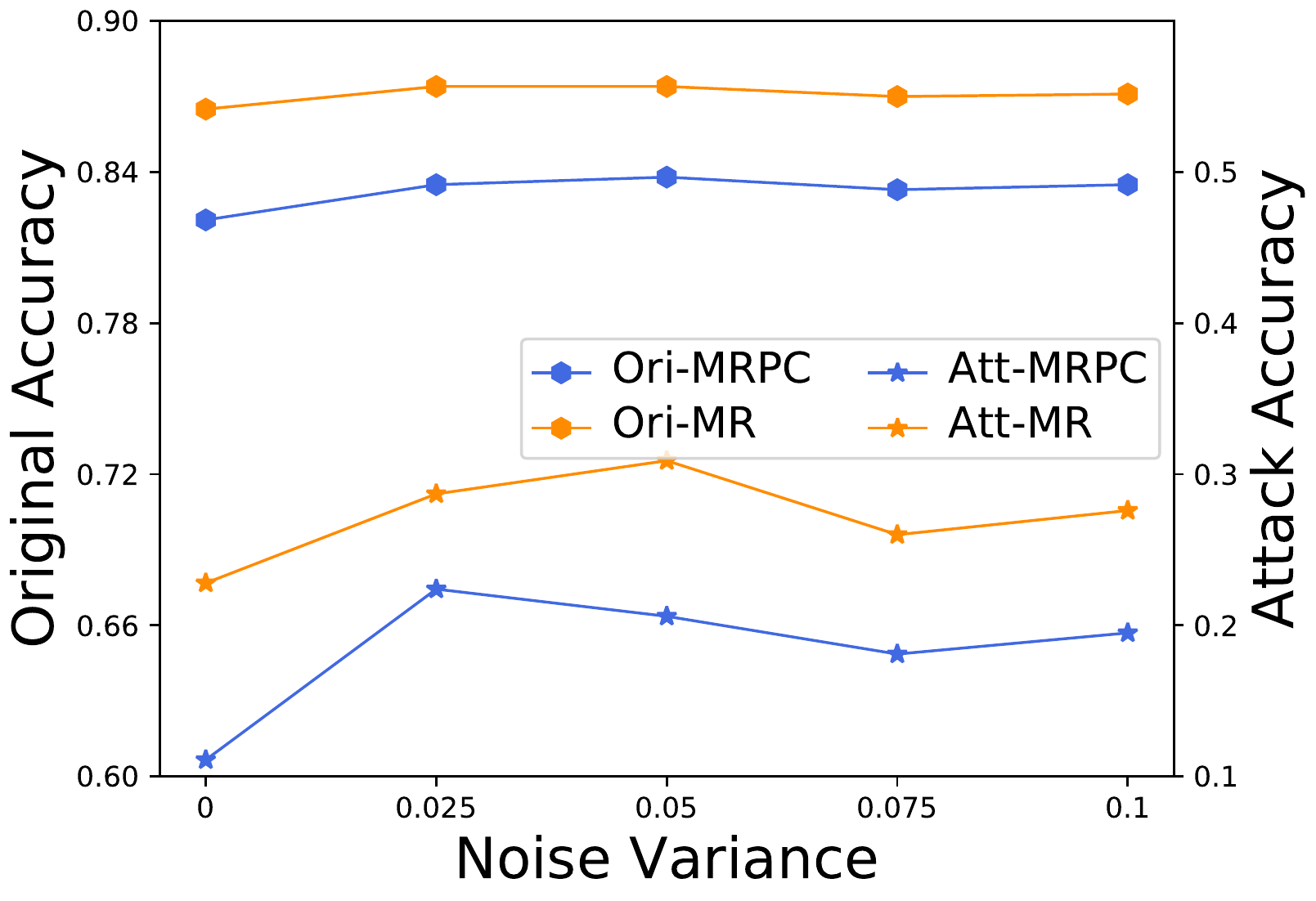}
\caption{Illustration of VDA.}
\end{figure}

\begin{figure}[t]
\centering
\includegraphics[width=\linewidth]{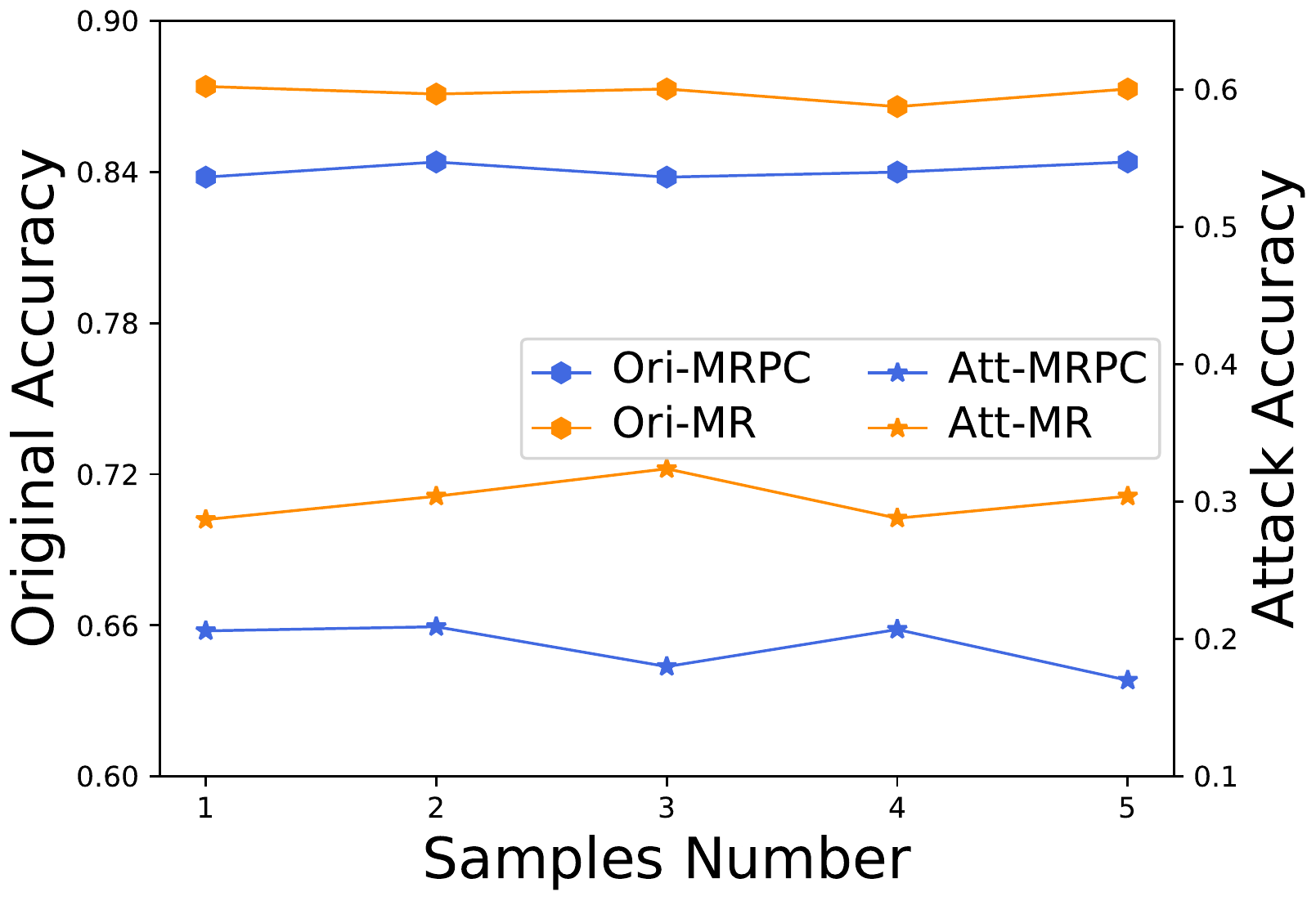}
\caption{Illustration of VDA.}
\end{figure}}

\begin{figure}[t!]
    \centering
    \begin{subfigure}[b]{0.49\linewidth}
        \centering
        \includegraphics[width=\textwidth]{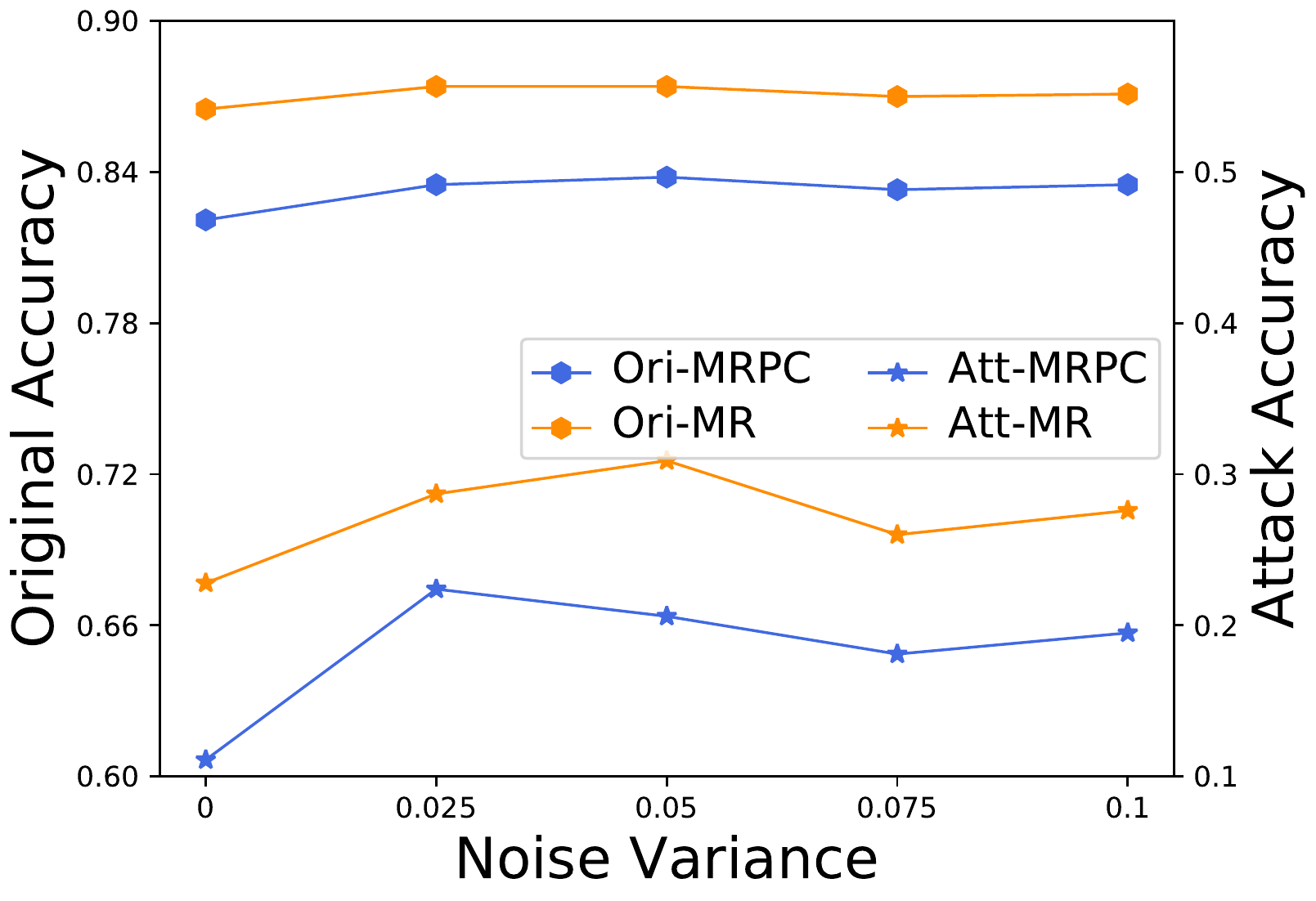}
        \caption{Noise variance $\eta$}
    \end{subfigure}
    \begin{subfigure}[b]{0.49\linewidth}
        \centering
        \includegraphics[width=\textwidth]{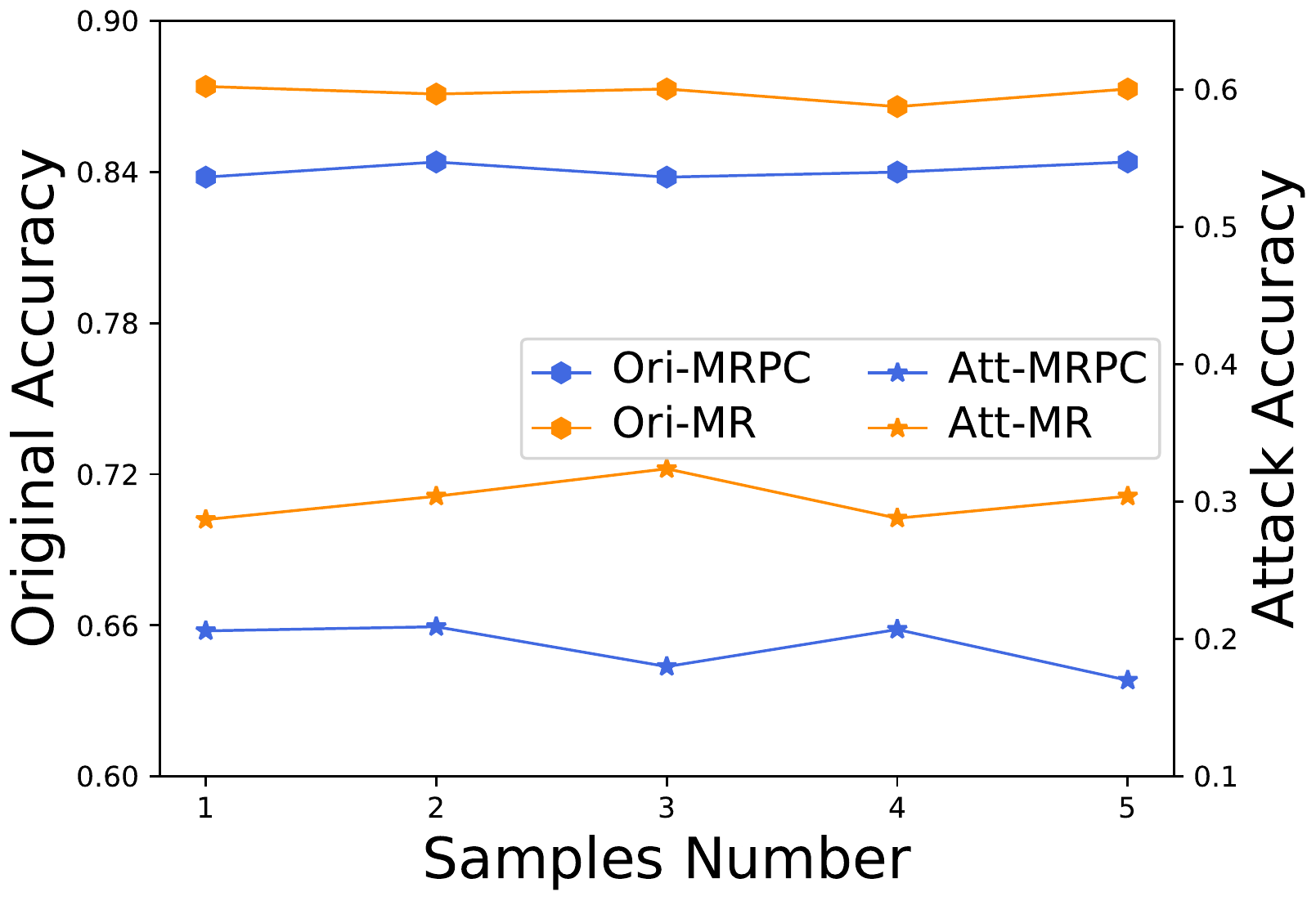}
        \caption{Sample number $n$}
    \end{subfigure}
    \caption{Performance comparison w.r.t. noise variance and sample number on the developed set of MR and MRPC datasets.}
    \vspace{-0.2cm}
\label{fig-param}
\end{figure}

Our framework includes a few parameters to tune. 
Here, we report the tuning results of two parameters on MR and MRPC datasets, \ie the variance of the Gaussian noise $\eta$ and the number of argumented virtual data.
We show the change curves of original accuracy and attack accuracy in Figure~\ref{fig-param}.
We can see that our model achieves the best performance when the variance is nearby 0.05.
It indicates that too small or too large noise may influence the quality of the augmented virtual data.
Besides, our model also achieves the best performance when the sampling number is nearby 3. It shows that augmenting 3 examples per sample is enough to improve the robustness.

\subsection{Performance Change during Regularizing Fine-tuning}
\begin{figure}[t!]
    \centering
    \begin{subfigure}[b]{0.49\linewidth}
        \centering
        \includegraphics[width=\textwidth]{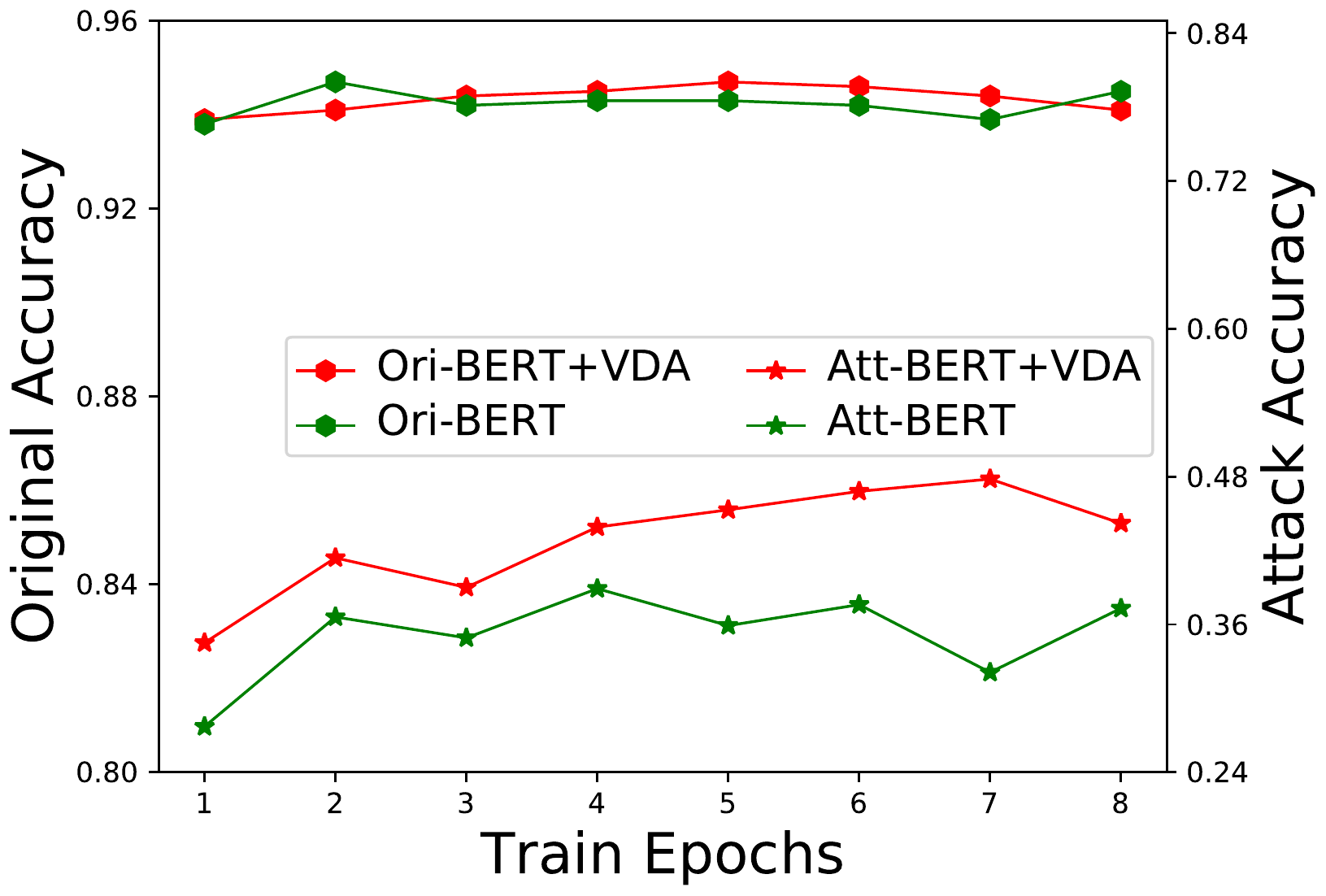}
        \caption{AG dataset}
    \end{subfigure}
    \begin{subfigure}[b]{0.49\linewidth}
        \centering
        \includegraphics[width=\textwidth]{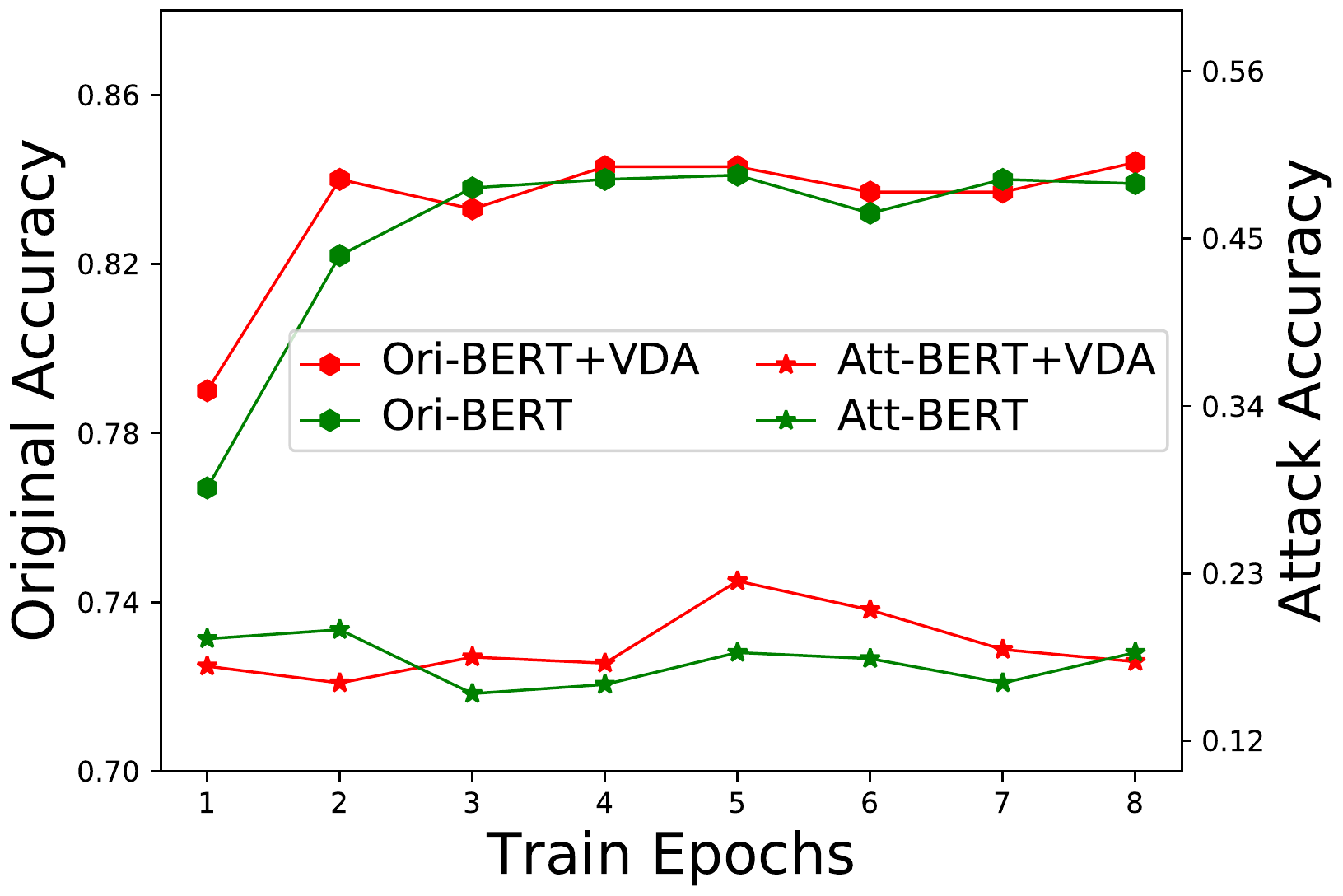}
        \caption{MRPC dataset}
    \end{subfigure}
    \caption{Performance comparison of BERT and BERT$_{VDA}$ w.r.t. training epochs on the developed set of AG and MRPC datasets.}
    \vspace{-0.2cm}
\label{fig-curve}
\end{figure}

In this part, we investigate how the accuracy and robustness change during regularizing fine-tuning with our VDA.
We conduct experiments on AG and MRPC datasets and report the original accuracy and attack accuracy metrics.
As shown in Figure~\ref{fig-curve}, the original and attack accuracy of the model can be improved with the increasing of training epochs.
When reaching the optimal point, the accuracy and robustness start to shock, and even decrease to some extent. The reason may be that the model has overfitted.
An interesting finding is that the optimal points of the original accuracy and attack accuracy are usually not the same one. A possible reason is that accuracy and robustness are not always consistent objectives for deep models.
Besides, we can see that after combined with our VDA, BERT is able to achieve a better optimal point with higher original and attack accuracy.
It indicates that VDA is an effective regularization approach for BERT.

\section{Conclusion}
In this work, we proposed the framework \emph{virtual data augmentation (VDA)}, for robustly fine-tuning pre-trained language models.
It is a general framework agnostic to downstream tasks, model architectures and learning strategies.
In VDA, we augmented new embeddings by making weighted aggregation on token embedding matrix according to a multinomial mixture distribution.
To construct the mixture distribution, we utilized a masked language model to generate the substitution probability for guaranteeing semantic consistency, and a Gaussian noise to provide diversity.
And we also adopted a regularized training strategy to further enhance the robustness.
Extensive experiments on six datasets have demonstrated that the proposed approach can effectively improve the robustness of various PLMs.

\section*{Acknowledgement}
We are thankful to Jinhao Jiang and Hui Wang for their supportive work and insightful suggestions.
This work was partially supported by the National Natural Science Foundation of China under Grant No. 61872369 and 61832017, Beijing Academy of Artificial Intelligence (BAAI) under Grant No. BAAI2020ZJ0301, Beijing Outstanding Young Scientist Program under Grant No. BJJWZYJH012019100020098, the Fundamental Research Funds for the Central Universities, the Research Funds of Renmin University
of China under Grant No.18XNLG22 and 19XNQ047, and Public Computing Cloud, Renmin University of China. Xin Zhao is the corresponding author.

\bibliography{anthology,custom}
\bibliographystyle{acl_natbib}

\end{document}